\def\year{2022}\relax
\documentclass[letterpaper]{article} % DO NOT CHANGE THIS
\usepackage{aaai22}  % DO NOT CHANGE THIS
\usepackage{times}  % DO NOT CHANGE THIS
\usepackage{helvet}  % DO NOT CHANGE THIS
\usepackage{courier}  % DO NOT CHANGE THIS
\usepackage[hyphens]{url}  % DO NOT CHANGE THIS
\usepackage{graphicx} % DO NOT CHANGE THIS
\usepackage{natbib}  % DO NOT CHANGE THIS AND DO NOT ADD ANY OPTIONS TO IT
\usepackage{caption} % DO NOT CHANGE THIS AND DO NOT ADD ANY OPTIONS TO IT
\DeclareCaptionStyle{ruled}{labelfont=normalfont,labelsep=colon,strut=off} % DO NOT CHANGE THIS
\usepackage{algorithm}
\usepackage{algorithmic}

\usepackage{url}
\usepackage{latexsym}
\usepackage{subfigure}
\usepackage{mathtools}
\usepackage{pgfplots}
\usepackage{tikz}
\usepackage{graphicx}
\usepackage{multirow}
\usepackage{amssymb}
\usepackage{bbding}

\newcommand{\newcite}[1]{\citeauthor{#1}~(\citeyear{#1})}

\usepackage{newfloat}
\usepackage{listings}
\floatstyle{ruled}
\newfloat{listing}{tb}{lst}{}
\floatname{listing}{Listing}
\title{Leveraging Advantages of Interactive and Non-Interactive Models for Vector-Based Cross-Lingual Information Retrieval}
\author{
    Linlong Xu, %\textsuperscript{\rm 1},
    Baosong Yang, %\textsuperscript{\rm 1},
    Xiaoyu Lv, %\textsuperscript{\rm 1},
    Tianchi Bi, %\textsuperscript{\rm 1},
    Dayiheng Liu, %\textsuperscript{\rm 1},
    Haibo Zhang %\textsuperscript{\rm 1}
}

\affiliations{
    Alibaba Group\\
    \{linlong.xll, yangbaosong.ybs, anzhi.lxy, tianchi.btc, liudayiheng.ldyh, zhanhui.zhb\}@alibaba-inc.com
}

\usepackage{bibentry}
\begin{document}

\maketitle

\begin{abstract}
Interactive and non-interactive model are the two  de-facto standard frameworks in vector-based cross-lingual information retrieval (V-CLIR), which embed queries and documents in synchronous and asynchronous fashions, respectively. 
From the retrieval accuracy and computational efficiency perspectives, each model has its own superiority and shortcoming.  
In this paper, we propose a novel framework to leverage the advantages of these two paradigms. Concretely, we introduce semi-interactive mechanism, which builds our model upon non-interactive architecture but encodes each document together with its associated multilingual queries. Accordingly, cross-lingual features can be better learned like an interactive model. 
Besides, we further transfer knowledge from a well-trained interactive model to ours by reusing its word embeddings and adopting knowledge distillation. 
Our model is initialized from a multilingual pre-trained language model M-BERT, and evaluated on two open-resource CLIR datasets derived from Wikipedia and an in-house dataset collected from a real-world search engine. 
Extensive analyses reveal that our methods significantly boost the retrieval accuracy while maintaining the computational efficiency.
%\footnote{Our codes will be released upon the acceptance of this paper.}
\footnote{Our codes are released on https://github.com/wutangA/Semi-Interactive-CLIR.}
\end{abstract}

\section{Introduction}
%1.传统检索方法在CLIR不适用，解决方法是语义匹配，但是跨语言的语言模型构建成为瓶颈
%Contrary to the 
Conventional cross-lingual information retrieval (CLIR) system mainly separates two stages, i.e., query translation and monolingual information retrieval~\cite{sunclireval,DBLP:journals/csur/ZhouTBWA12,DBLP:journals/corr/abs-1912-12481,DBLP:conf/aaai/LiLWBLL0Y20,DBLP:conf/coling/YaoYZCL20}. On the contrary, vector-based CLIR (V-CLIR) \cite{vulic2015monolingual,yarmohammadi-etal-2019-robust} employs neural networks to encode input texts and score their associated similarity with end-to-end training, thus avoiding the problem of error propagation in translation-based systems~\cite{wu2010study,DBLP:conf/clef/BoscaD10}. 
%\cite{huang2013learning} first introduce  deep structured semantic model (DSSM) 
%Considering the neural-based retrieval models, 

A widely used neural-based retrieval framework is deep relevance matching model \citep[DRMM,][]{guo2016deep}, which is built upon a unified semantic matching model to generate embeddings for queries and documents. In this paradigm, features of query and document are directly interacted at the representation learning time (being called as interactive mechanism, as shown in Figure 1 (a)), leading to promising results in a variety of information retrieval tasks \cite{mikolov2013distributed,bojanowski2017enriching}.  
However, in a real-world search engine, a user query has to match with millions of documents in database. When performing the interactive framework, for each query, massive real-time computations have to be carried out for obtaining joint representations.  

 Towards approaching this problem, existing V-CLIR systems are mainly built with a non-interactive architecture~\cite{huang2013learning,reimers2019sentence,lu2020twinbert}, as shown in Figure 1 (b). This model encodes queries and documents asynchronously, making the offline learning of document representations available. %achieve the offline modeling and storage with respect to document. % are modeled offline and stored in cache. 
 In this way, each query only requires to be encoded once during online inference. 
 Despite its superiority on computational efficiency, such kind of architecture lacks in feature fusion among queries and documents, further raising the difficulty of cross-lingual semantic matching \cite{reimers2019sentence,lu2020twinbert}. 
 It can be said that how to build a fast yet high-quality deep semantic matching model to enhance V-CLIR still remains a great challenge.
 
 \begin{figure*}[t]
\centering
\subfigure[Interactive Mechanism]
    {
        \begin{minipage}[t]{0.27\textwidth}
        \centering         %子图居中
        \includegraphics[width=0.75\textwidth ]{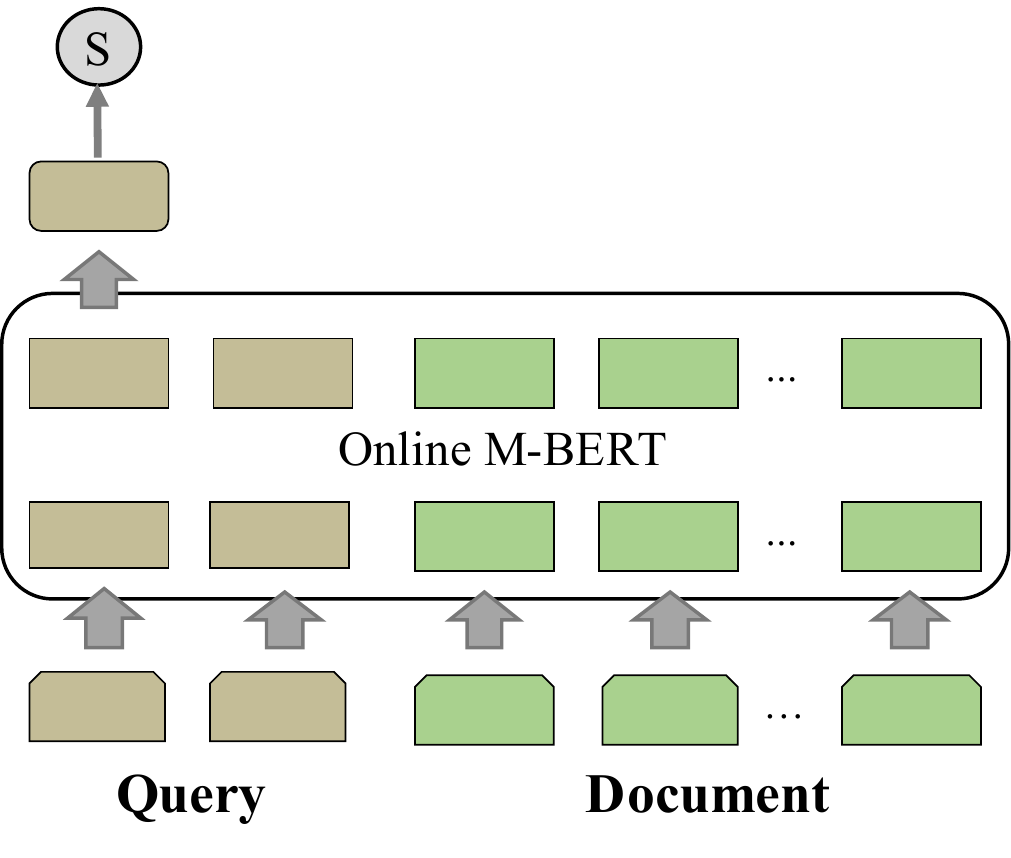}
        \end{minipage}%
    }
\subfigure[Non-Interactive Mechanism]
    {
        \begin{minipage}[t]{0.28\textwidth}
        \centering         %子图居中
        \includegraphics[width=0.76\textwidth]{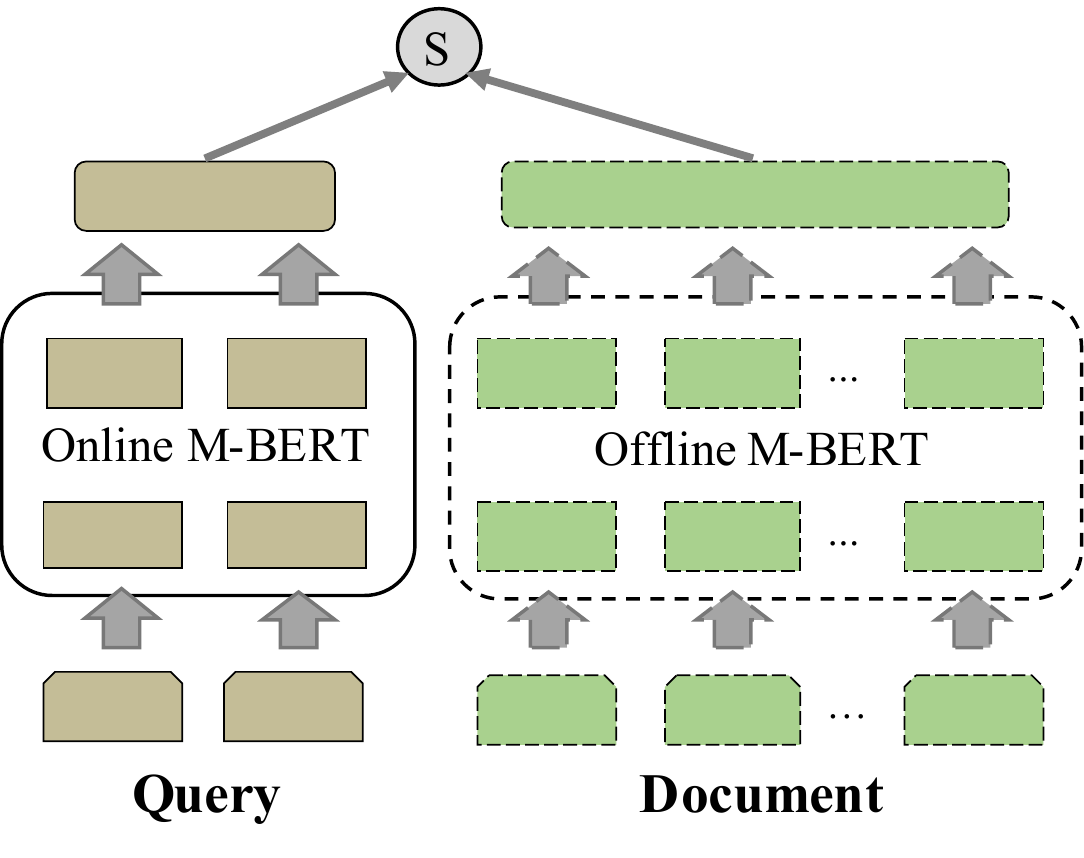}
        \end{minipage}%
    }
\subfigure[Ours: Semi-Interactive Mechanism]
    {
        \begin{minipage}[t]{0.40\textwidth}
        \flushleft         %子图居中
        \includegraphics[width=0.96\textwidth]{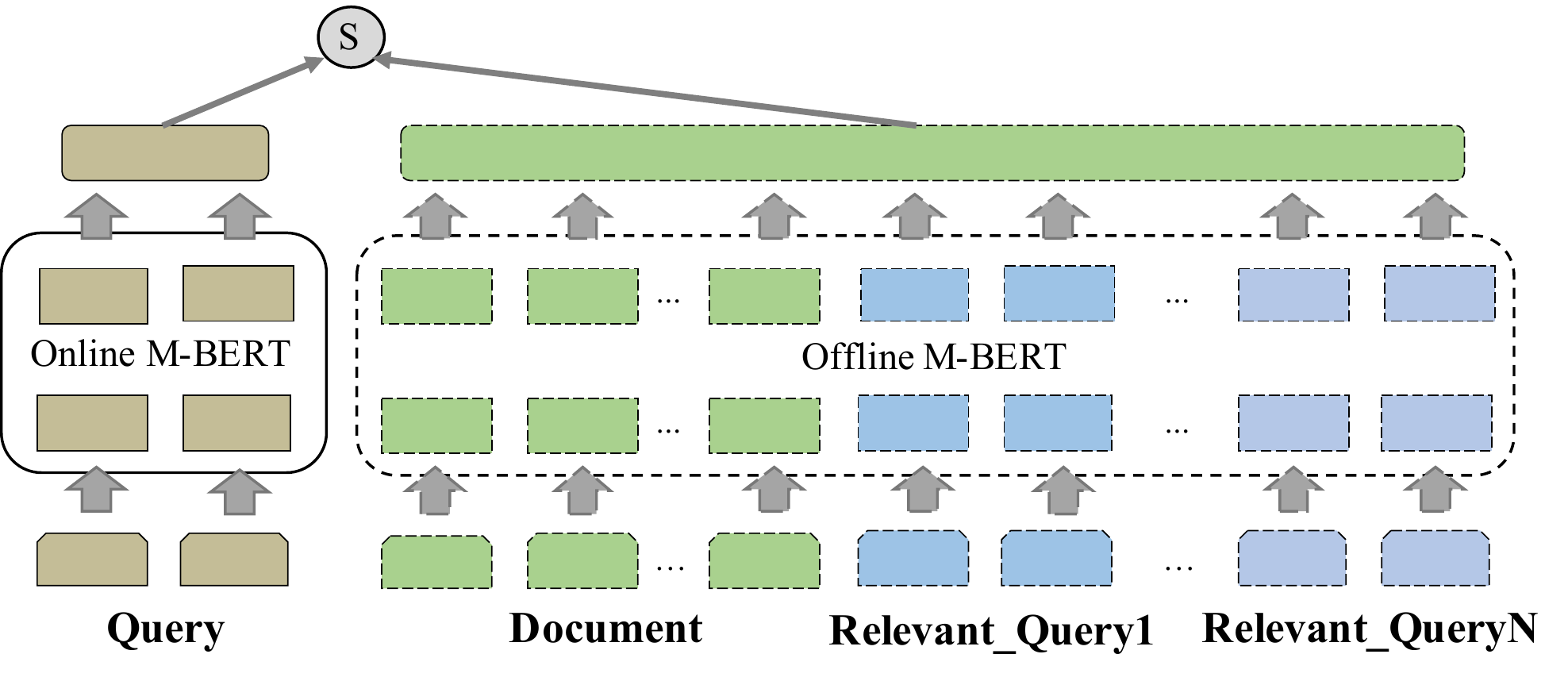}
        \end{minipage}%
    }
\caption{Illustration of a) interactive mechanism which encodes inputs jointly, b) non-interactive architecture that learns representations of query and document separately, and 3) the proposed semi-interactive mechanism in which the relevant multilingual queries are incorporated into document representation thus narrowing the cross-lingual semantic latent spaces. }% of query and document. } %As seen, interactive model encodes query and document jointly, while non-interactive learns representations of query and document separately. Our model incorporates associated multilingual queries into document representation for narrowing the semantic latent space.}% A popular deep structure semantic matching model used in the mono-lingual IR systems which is built upon a pre-trained language model -- BERT~\cite{devlin2018bert}. For efficiency, the query and documents are encoded separately regardless of the feature interaction among inputs.}
\label{fig:models}
\end{figure*}
 A natural question arises: \textit{Is it possible to leverage the advantages of interactive and non-interactive mechanisms simultaneously?} 
 In this paper, we propose a novel model named semi-interactive mechanism, as visualized in Figure 1 (c).  
 Concretely, our model is built upon a non-interactive framework to maintain the  computational efficiency. In order to narrow the distance between the latent spaces of cross-lingual queries and documents, we fuse the information of a document with its associated multilingual queries, which can be either ready-made or collected from the search log. %have the same or similar meaning in different languages and could be collected by several manners as described in Section~\ref{sec:sim} for detail.
 The representation of document  therefore contains rich multilingual contexts. %enrich the document representation with multi-lingual features. 
 %Besides, we set a well-trained interactive model as teacher and transfer its knowledge to our model by reusing word embeddings and adopting knowledge distillation at the training time \cite{sanh2019distilbert,sun2019patient}. 
 To further leverage the advantage of interactive mechanism, we set a well-trained interactive model as teacher and transfer its knowledge to our model by reusing word embeddings and adopting knowledge distillation at the training time \cite{sanh2019distilbert,sun2019patient}. 
 With these enhancements, the proposed model is able to preserve the online inference speed like a non-interactive model, in the meanwhile, fill the quality gap to its interactive counterpart.

Following the common settings \cite{yilmaz2019applying,jiang2020cross}, we build our approach upon a multilingual pre-trained language model -- M-BERT~\cite{devlin2018bert}. % to preliminarily map multilingual texts into the same latent space. 
%In order to evaluate the effectiveness of our model, 
%We train our model on WikiCLIR \cite{sasaki2018cross}, a large-scale CLIR dataset for Russian-to-English, Spanish-to-English, Portuguese-to-English, French-to-English and Arabic-to-English document retrieval. 
%Then evaluations are made on both WikiCLIR and CLIRMatrix MULTI-8{\cite{sun2020clirmatrix}}. 
The training set is constructed from a large-scale CLIR dataset -- WikiCLIR \cite{sasaki2018cross},  for Russian-to-English, Spanish-to-English, Portuguese-to-English, French-to-English and Arabic-to-English document retrieval. 
%Our model are examined on both of relevance tasks collected from WikiCLIR and retrieval tasks 
Our model is examined on both semantic similarity tasks collected from WikiCLIR and document search tasks 
provided in  MULTI-8 \cite{sun2020clirmatrix}. 
%Each training set consists of around 1 million to 5 million training samples collected from the Large-Scale CLIR Dataset \cite{sasaki2018cross}. For each retrieval direction, around 60K test examples are extracted with manually labeling. 
Empirical results demonstrate that the proposed semi-interactive mechanism and knowledge transferring progressively improve the retrieval accuracy over non-interactive baseline and marginally drop the inference speed. 
In addition, we conduct experiments on a real-world search engine. Our model still significantly outperforms non-interactive based approach.
Our visualization of representation distribution reveal that the proposed model can approximately transform both query and document representations to a standard orthogonal basis~\cite{DBLP:journals/corr/abs-2103-15316}, leading to more discriminative representations for similarity computation. 
%Furthermore, \cite{DBLP:journals/corr/abs-2103-15316,DBLP:conf/iclr/GaoHTQWL19} explore the reason for the poor performance of BERT-based sentence embedding in similarity matching tasks, i.e., it is not in a standard orthogonal basis, we project the query and document embeddings into 3-dimensional space using principal component analysis \citep[PCA,][]{abdi2010principal} for visual comparison. 
%Furthermore, we project the query and document representations into a 3-dimensional space using principal component analysis \citep[PCA,][]{abdi2010principal} for visual comparison. The results reveal that our model can approximately transform both query and document representations to a standard orthogonal basis, which is consistent with previous findings \cite{DBLP:journals/corr/abs-2103-15316}. That means the proposed methods exactly learn better query and document representations for similarity computation. 
%We propose contextualized representations for CLIR. Specifically, we first pre-train a large-scale contextualized language model to map syntactic and semantic representations of different languages into a same latent space.Then, a novel semi-interactive mechanism is proposed to encode a document together with its associated multilingual queries, which overcomes the lack of interaction between queries and documents. Experiments  show that the proposed methods progressively improve the retrieval accuracy of CLIR and eventually surpasses the translation-based baseline. 

%To summarize, 
Major contributions of our work are three-fold:
\begin{itemize}
    \item We introduce a novel semi-interactive mechanism, offering V-CLIR model ability to combine the advantages of interactive and non-interactive paradigms.  
    %1) To map the representations of inputs into the shared latent space, we propose to contextualize these representations by 
    %pretraining an in-domain cross-lingual language model and 
    %exploiting a semi-interactive mechanism. 
    \item We propose to transfer knowledge from a well-trained interactive model, thus boosting the retrieval  quality of a non-interactive model. %collect CLIR benchmark in E-Commerce scenario and make them publicly available, which may contribute to the subsequent researches in CLIR community. 
   \item %Extensive analyses indicate the effectiveness of our work and verify that our model can exactly learn better query and document representations with the help of the proposed methods.  
   %Extensive analyses indicate the universal-effectiveness of our work and verify that our model is able to learn similar representation distribution for multilingual texts. % into the same latent space.
   Extensive analyses indicate the universal-effectiveness of our work and verify that our model is able to learn better representation for multilingual texts.
   
   %Experimental results show that the proposed method significantly outperforms non-interactive and knowledge distillation baselines on CLIR tasks. Our study provides an appealing alternative to abandon language identification and translation procedures in existing CLIR systems.
\end{itemize}

\section{Preliminary} 
%In this section, we introduce related work and background on CLIR. 
\subsection{Cross-Lingual Information Retrieval}
%The research on cross-language information retrieval can be traced back to the publication of Mr. G. Salton's Experiments in multilingual information retrieval in 1973. The research at that time was mainly for international online retrieval. Because the retrieval system is not popular, people's demand for network information is not strong. 
%With the development of Internet and information globalization, 
CLIR has attracted increasing interests over the past decades. The main challenge is to bridge the language discrepancy between user queries and documents~\cite{nie2010cross}. Most of existing approaches use translation based technique, which first calls a machine translation (MT) model to translate either queries or documents into the same language \cite{levow2005dictionary,da2017cross}, the translation results are then passed to a monolingual IR system. However, such kind of framework requires huge amounts of parallel data, which is scarce for many language pairs and domains in low-resource. In addition, the search results are directly affected by translation accuracy. 

With this point, researches start to explore vector-based IR methods \cite{DBLP:conf/kdd/NigamSMLDSTGY19,DBLP:journals/ftir/MitraC18,tai2002information,mihalcea2006corpus,bi2020constraint,yao2020exploiting} without the need to build MT systems. In this context, queries and documents are represented as vectors for similarity computation. 
%Cross language information retrieval without machine translation\cite{sasaki2018cross}, introduce a large-scale dataset derived from Wikipedia to support CLIR research in 25 languages. The validity of the data is verified. But, the twin-tower model is not optimized and the pre-training is not used.

%Therefore, translation between two or more languages has an important impact on the performance of cross-language retrieval. Commonly used cross-language information retrieval technology mainly consists of two ways: the first is query translation, which translates the user query language into a target language (such as English), and then uses the target language for subsequent retrieval. However, this method relies heavily on the accuracy of translation; the second is document translation, which does not translate query, directly uses the language of user query to establish document index for retrieval, and finally translates into target language when displaying. However, this method requires different indexes for different languages, especially in cross-border e-commerce scenarios, which will consume huge storage resources.
\subsection{Vector-Based Retrieval Models}
%Traditional IR methods based on text rely on a literal matching relationship. 
%For the text-based retrieval,  the contents in each document are used to perform word-based indexing. The similarity score for each document with respect to a query is computed by Okapi BM25 \cite{robertson1995okapi} or TF-IDF \cite{ramos2003using}.
%Nevertheless, the fuzzy searches in real-world scenario such as misspelling and code-switching make traditional methods fail to be well handled \cite{landauer1997solution}. 
%With this point, researches start to explore vector-based IR methods \cite{tai2002information,mihalcea2006corpus}. In this context, user queries and documents are represented as vectors for similarity computation. 
%From the model architecture perspective, 
Vector-based IR model can be divided into two directions: interactive and non-interactive models. 
\subsubsection{Interactive Mechanism}
Deep relevance matching model~\citep[DRMM,][]{guo2016deep} and kernel-based neural ranking model ~\citep[KNRM,]{xiong2017end}
%Deep structured semantic models (DSSM)~\cite{huang2013learning} and its convolutional variant CDSSM \cite{shen2014learning} 
are two representatives under interactive framework. % that exploit deep neural networks for semantic matching. 
These approaches concatenate and feed query-document pairs into a unified deep neural encoder to learn their representations for semantic matching, as shown in Figure~\ref{fig:models}(a). Accordingly, features in query and document are interacted and fused, resulting in promising quality on a variety of IR tasks \cite{mcdonald2018deep}.  Recent studies further initialize the encoder with large-scale pre-trained language model, such as BERT~\cite{devlin2018bert}, which brings better model abilities on representation learning and relevance scoring \cite{jiang2020cross}. 
 %widely used architecture is deep structured semantic models (DSSM) \cite{huang2013learning} which exploits neural networks to learn the representations of queries and documents, respectively (Figure~\ref{fig:models}(a)). 
Specifically, given the input token embeddings of a query $\textbf{Q}$ and a document $\textbf{D}$, the interactive model first encodes them into hidden states: %in non-interactive mechanism can be represented as:
%\begin{equation}
%\textbf{Q}=\{[CLS],Q_{1},…,Q_{{N}},[SEP]\}
%\end{equation}
%\begin{equation}
%\begin{aligned}
%\textbf{D}=\{[CLS],D_{1},…,D_{{M}},[SEP]\}
%\end{aligned}
%\end{equation}
%where $[CLS]$ and $[SEP]$ are assigned as the beginning and ending symbol, respectively.
\begin{equation}
\textbf{H}=\mathrm{Encoder}([\textbf{Q},\textbf{D}])
\end{equation} 
where $\mathrm{Encoder}(\cdot)$ indicates the BERT model, $[\cdot]$ represents the concatenation of inputs. Then, a non-linear active function is adopted to calculate the relevance score $\textbf{s}$ over the mean of the output hidden states $\overline{\textbf{H}}$:
\begin{equation}
\textbf{s} =\mathrm{sigmoid}(\overline{\textbf{H}})
\end{equation} 
%where $[CLS]$ and $[SEP]$ are assigned as the beginning and ending symbol, respectively.
 %most of vector-based model architectures are non-interactive . 
 
%The main problem of such interaction models, except that the performance of a single inference time is affected by the model structure, the more important problem is that, for instance, when we put a query
%and want to retrieve the 10 most relevant documents from 1 million documents, we need to calculate the similarity between query and documents 1 million times separately .It is a catastrophic problem for a real-time IR system.
 
\subsubsection{Non-Interactive Mechanism} 
%Deep structured semantic models (DSSM)~\cite{huang2013learning} and its convolutional variant CDSSM \cite{shen2014learning} have pioneered the context of using deep neural networks for semantic matching in non-interactive architecture (Figure~\ref{fig:models}(b)). 
Contrary to the synchronous encoding procedure in interactive architecture, the representations of queries and documents in non-interactive architecture (Figure~\ref{fig:models}(b)) are encoded asynchronously~\cite{huang2013learning,shen2014learning,reimers2019sentence,lu2020twinbert}. % In this way, even using pre-trained language models, document representations can be pre-computed offline thus speeding up the online retrieving\cite{lu2020twinbert,reimers2019sentence}. 
%A widely used architecture is deep structured semantic models (DSSM) \cite{huang2013learning} which exploits neural networks to learn the representations of queries and documents, respectively. 
%The input token embeddings of query $Q$ and document $D$ in non-interactive mechanism can be represented as:
It employs two encoders for query and document representation learning, respectively. % contains two M-BERT models to independently encode the query and each retrieved document. %M-BERT exploits the advanced Transformer \cite{vaswani2017attention} architecture, which is highly parallelized and has shown standout model capability on natural language processing~\cite{devlin2018bert}. 
Formally, the representations of query and document are calculated in a separate manner:
\begin{equation}
\label{eq:doc_enc}
\textbf{H}_Q=\mathrm{Encoder}_Q(\textbf{Q}), ~~~~ \textbf{H}_D=\mathrm{Encoder}_D(\textbf{D})
\end{equation}
%\begin{equation}
%\textbf{H}_D=\mathrm{Encoder}_D(\textbf{D})
%\end{equation} 
A cosine function is assigned for the similarity calculation: % of similarity between $\overline{\textbf{H}_Q}$ and $\overline{\textbf{H}_D}$:%, which is the dot-product of the normalization with respect to \textbf{Q} and \textbf{D}:
\begin{equation}
\textbf{s} =\cos (\overline{\textbf{H}}_Q,\overline{\textbf{H}}_D) = \frac{\overline{\textbf{H}}_Q \cdot \overline{\textbf{H}}_D}{||\overline{\textbf{H}}_Q|| \cdot ||\overline{\textbf{H}}_D||}\label{con:modelsim}
\end{equation}
\subsubsection{Training}
The parameters of both interactive and non-interactive models are optimized by minimizing the cross-entropy loss over the training set:

\begin{equation}
\textbf{L}_{cross} = -\sum_{\textbf{Q},\textbf{D}}\textbf{y}\mathrm{log}(\textbf{s}) + (1-\textbf{y})\mathrm{log}(1-\textbf{s}) \label{con:modelloss}
\end{equation}
where $\textbf{y}\in \{0, 1\}$ denotes the groundtruth relevance label of $\textbf{Q}$ and $\textbf{D}$. %, and $\mathrm{s=s}(\textbf{Q},\textbf{D})$ is the similarity as defined in equation (\ref{con:modelsim}).

\section{Methodology}

\subsection{Motivation} 
%For each $\textbf{Q}$, interactive model has to recompute the joint representation $\overline{\textbf{H}}$ millions times for all the documents in database.  This is unavailable in a real-world search engine, as IR systems have to be real time \cite{zhang2015information}.  
To find the most similar document of $\textbf{Q}$ in a collection of millions documents, interactive model has to recompute the joint representation $\overline{\textbf{H}}$ millions times for all the documents.
%This is unavailable in a real-world search engine, as IR systems have to be real time \cite{zhang2015information}. 
This massive computational overhead makes it unsuitable for real-word search engine \cite{zhang2015information}.
The superiority of non-interactive architecture lies in the fast online inference. Nevertheless, the lack of interaction between $\textbf{Q}$ and $\textbf{D}$ may cause biases between latent vectors, raising the difficulty on semantic matching. In particular, $\textbf{Q}$ and $\textbf{D}$ in different languages further deteriorate this problem, resulting in unsatisfied retrieval quality. 

Accordingly, our goal is to leverage the advantages of interactive and non-interactive models. 
%Considering cross-lingual scenarios, the challenge of vector-based approach is the lack of a solution to built cross-lingual representations that share the same latent space, especially when the vector-based approach requires the model to adopt a non-interactive structure.
In this work, we handle this problem from two aspects: 
\begin{itemize}
    % \item Partially inspired by the success of pre-trained model in monolingual IR community~\cite{reimers2019sentence,lu2020twinbert}, we  pre-train a cross-lingual language model using large-scale queries and documents in different languages,
    % %Recent studies on multilingual unsupervised pretraining  demonstrates 
    % which demonstrates the potential effectiveness on building linguistic connections across languages %\cite{devlin2018bert,conneau2019unsupervised}.%produces a representation that seems to generalize well across language..
    \item  
    %In order to maintain the inference speed, 
    From the model perspective, we propose semi-interactive module (Figure~\ref{fig:models}(c)) to encode the representation of a document with its relevant multilingual queries. In this way, multilingual features can be supplemented  into the document representation. %Accordingly, the document representation contains contexts with respect to relevance multi-lingual quires, in the meanwhile, maintaining the inference speed as same as non-interactive models.
    %~\cite{huang2013learning}. 
    \item From the optimization perspective, we 
    introduce knowledge transfer methods, in which a well-trained interactive model is regarded as teacher. We reuse its word embeddings and distill its knowledge to our model. %thus leveraging the advantage of interactive model.
    %knowledge transfer 
    %In order to leverage the advantage of interactive model on semantic matching, we further transfer knowledge from a well-trained interactive model to ours.
    
\end{itemize}

\subsection{Semi-Interactive Mechanism}
\label{sec:sim}
One principle of our model is to maintain the inference speed, in the meanwhile, improve retrieval quality. To this end, we build the proposed model upon a non-interactive architecture. A natural solution to alleviate the discrepancy between latent spaces of query and document is to enhance the document representations with cross-lingual features. %Since there exists a large  number of queries, of which representations have to be modeled online, we pay our attention on the document side for computational efficiency.  
Specifically, we propose to provide  multilingual contexts for the document encoder. In this way, cross-lingual properties are incorporated at the representation learning time, thus producing more informative representations. More importantly, since each document and its associated contexts can be encoded offline, the computational efficiency is still preserved. %such kind of manner can be still applied offline. 
We serve the multilingual keywords or queries that is relevant to the document as its multilingual contexts.  
%modeling query representation has to   
%Our model is built based on a non-interactive architecture. %The differences are described below:

\begin{figure*}[t]
\centering
\includegraphics[width=0.9\textwidth]{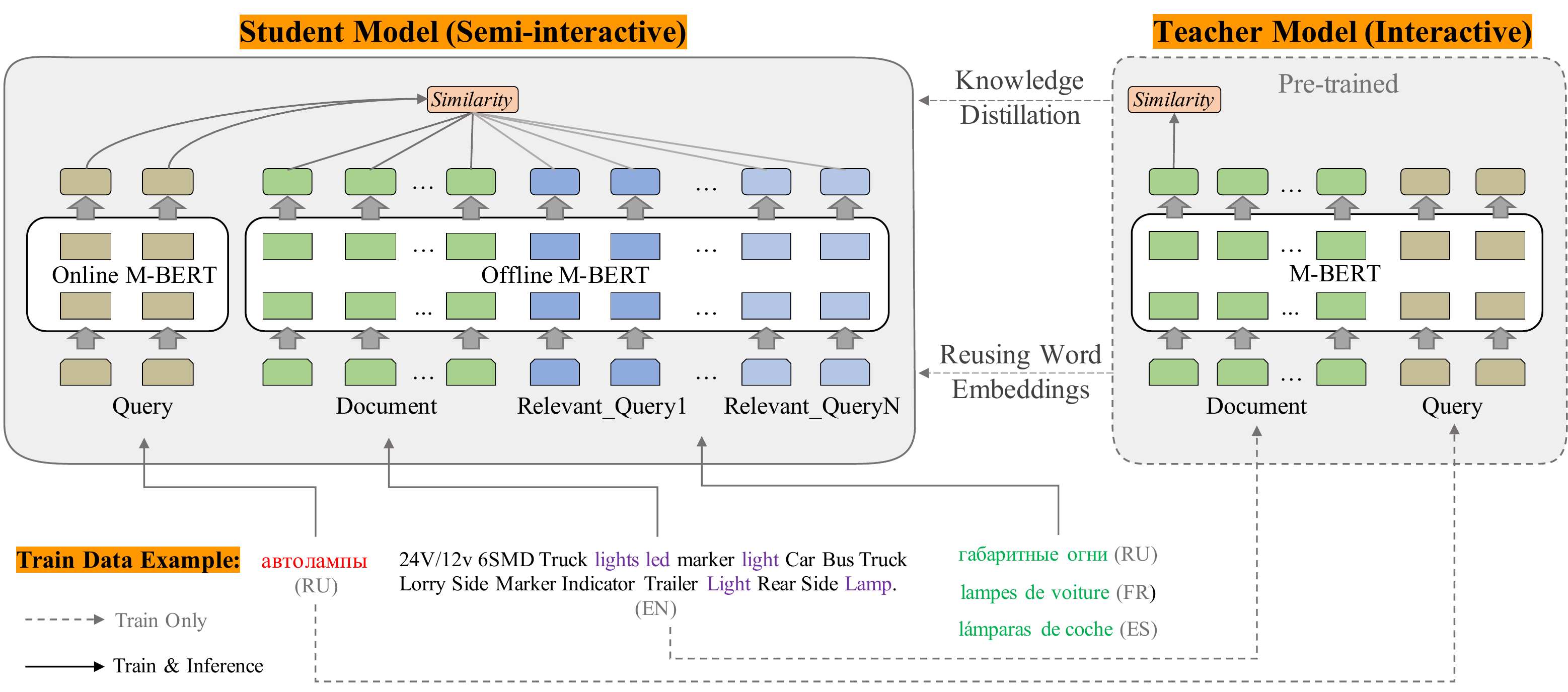} % Reduce the figure size so that it is slightly narrower than the column.
\caption{Illustration of our model architecture. We cast the proposed semi-interactive approach as a student model in which each document is supplemented with its relevant multilingual queries. We further transfer knowledge from the well-trained interactive model to semi-interactive one by: 1) reusing word embeddings, and 2) distilling knowledge.
%The encoder is initialized by a  multi-lingual language model (M-BERT). In order to further closer the query and document embeddings, we propose a semi-interactive approach which encodes a document with its related top K multi-lingual queries.  
%Finally, the similarity in terms of query and document is calculated by a cosine function. 
}
%probability distribution of the languages with respect to input query is predicted through the output layer.}
\label{fig:semi_model}
\end{figure*}

%模型输入，query，document，topkQuery，
%\paragraph{Document Input Representation}
%In our model, the query and document are decoupled and encoded separately. 
As shown in Figure~\ref{fig:semi_model}, given $N$ relevant queries $\{\textbf{Q}^r_1,…,\textbf{Q}^r_N\}$ for the document $\textbf{D}$, where $\textbf{Q}^r_i$ ($i\in \{1, ...,N\}$) denotes the $i$-th relevant query, the document encoder produces joint representations $\textbf{H}_D$ by modifying Equation \ref{eq:doc_enc} as following:
\begin{equation}
\textbf{H}_D=\mathrm{Encoder}_D([\textbf{D},\textbf{Q}^r_1,…,\textbf{Q}^r_N])
\end{equation} 
In this paper, we set $N=3$ as default according to ablation study as Section \ref{abs} described.
\iffalse
By mining relevant query \textsc{RelQ}, which is described later in the semi-interactive mechanism part, document and associated queries are jointly fed to document encoder.
%We assign $[CLS]$ and $[SEP]$ as the beginning and ending symbol, respectively. 
%In semi-interactive mechanism, document and associated queries are jointly fed to our encoder, which are segmented by $[SEP]$. % the blank token,[E]as the special token indicating the end ofthe sentence, and[U]as the unknown tokens
%The input token embeddings of query are the same with non-interactive two, while 
The input token embeddings of document can be represented as:
\begin{equation}
\begin{aligned}
\textbf{Tok}_\textbf{\textsc{D-RelQ}} = \textbf{Tok}_\textbf{D} + \textbf{Tok}_\textbf{\textsc{Q-Rel}} =\{[CLS],Tok_{d_1},…,Tok_{{d_N}},[SEP],\\
Tok_{q_1},…,Tok_{q_k},[SEP]\}
\end{aligned}
\end{equation}
\paragraph{Document Output Representation}
Formally, the output of the document M-BERT is calculated as:
\begin{equation}
\textbf{D}=\mathrm{Encoder}(\textbf{Tok}_\textbf{\textsc{D-RelQ}})
\end{equation} 

In order to further closer the query and documents,
a novel semi-interactive mechanism is proposed to encode a document together with its associated multilingual queries. Accordingly, the document representation can be enhanced with the cross-lingual contextual information. 
\fi

\subsubsection{Collecting Relevant Queries}
\label{sec:topkm}
%We give an example of relevant queries for a document, as shown in Figure~\ref{fig:topkcase}. 
Several potential manners can collect relevant multilingual queries: 
\begin{itemize}
\item{\textbf{Ready-Made Queries}.} 
Several documents have already provided multilingual summarization or keywords. For example, the dataset WikiCLIR~\cite{sasaki2018cross} collects large amount of pages of Wikipedia, each of which links with multilingual relevant keywords. For simplification, we can directly collect $N$ ready-made relevant queries labeled as ``relevant'' for each document.  %Specifically, for each document, we randomly extract N relevant queries as its multilingual contexts. 
\item{\textbf{Mining-Based Queries}.} For an existing cross-lingual search engine (no matter translation-based or vector-based), the relevant queries for each document can be extracted according to its clickthrough data. 
%We can utilize the relationships of users and queries, as well as queries and documents for the construction of two bipartite graphs containing three types of vertices u, q, d. 
%which reflects users' interests and
The click behavior reflects users' interests and the latent semantic relationships between the input queries and the clicked documents \cite{radlinski2008does,ma2008learning}. We can collect Top $N$ queries according to the click-through rate (CTR).  
%\item{\textbf{Generation-Based Methods}.} Considering a initial model, 
\end{itemize}
A special circumstances is that there is neither off-the-rack multilingual relevant queries, nor search log. In this paper, we do not take attention into this case, because a vector-based CLIR system even can not be well-trained without large-scale cross-lingual annotated data, letting alone considering the  trade-off between efficiency and quality. Despite of that, several few-shot learning approaches in IR contexts can be carried out to handle this problem, such as generating pseudo data using MT \cite{DBLP:conf/rep4nlp/ChidambaramYCYS19,DBLP:conf/acl/YangCAGLCAYTSSK20} and  searching associated queries using a preliminary retrieval system \cite{DBLP:conf/sigir/LitschkoGPV18}.

\subsection{Knowledge Transfer}
Another promising direction is to transfer knowledge from the well-trained interactive model to the non-interactive one. As interactive architecture benefits to the representation learning and relevance scoring, we treat it as a teacher to assist the training of non-interactive models. Our model profits from the teacher model in two aspects: 1) initializing word embeddings of our model with that of teacher model and 2) distilling prediction distribution of teacher model to ours. 
\subsubsection{Reusing Word Embeddings} The distribution of word embeddings immediately affects the final representations, therefore plays a critical role in neural-based natural language understanding tasks \cite{levy2014dependency,kusner2015word}. As cross-lingual queries and documents are jointly encoded by the interactive model,  word embeddings have relatively less discrepancy caused by distinct languages. As a result, we reuse the well-trained word embedding layer of the teacher model to initialize that of non-interactive models. We expect this can carry a better initialization for model parameters, thus contribute to the subsequent training procedure. % We use the well-trained interactive teacher model to initialize the embedding of semi-interactive student model, including query and document. 
\subsubsection{Knowledge Distillation} %In order to leverage the advantage of interactive model on semantic matching, 
%We further exploit knowledge distillation \cite{sanh2019distilbert,sun2019patient} to guide the non-interactive model to learn the relevance prediction from its teacher. 
Knowledge distillation approach \cite{DBLP:conf/nips/BaC14,DBLP:journals/corr/HintonVD15} enables the transfer of knowledge from a teacher model to  a student model, which is improved in the process. 
%Notice that all of these  distillation approaches either transfer of knowledge from a large model to a same smaller or a completely different structure model.  The question is: \textit{Is it possible to distill knowledge from the interactive model to similar semi-interactive model?}  
%这些蒸馏方法的问题：1）为了压缩模型，都是从大模型蒸馏到小模型； 
%But when the structure of the student and teacher models are quite different, the effect is often not good.
Here, we exploit knowledge distillation to guide both non-interactive and semi-interactive model to learn the probability distribution output by a well-trained interactive model. % and its parameters are fixed. % We implement knowledge distillation by modifying the loss function of student model.
Specifically, given the relevance probability  $\textbf{s}{'}$ predicted by the teacher model, the goal of distillation is to force our model to fit the probability distribution of the teacher: 
\begin{equation}
\textbf{L}_{distill} = -\sum_{\textbf{Q},\textbf{D}}\textbf{s}{'}\mathrm{log}(\textbf{s})% + (1-\textbf{s}{'})\mathrm{log}(1-\textbf{s}) %\label{con:modelloss}
\end{equation}
The final loss can be formally expressed as:
\begin{equation}
\textbf{L} = \alpha\textbf{L}_{cross} + (1-\alpha)\textbf{L}_{distill}
\end{equation} 
where $\textbf{L}_{cross}$ is the conventional cross-entropy loss as defined in Equation \ref{con:modelloss}. $\alpha$ denotes a factor to balance the two loss, which is set to 0.7 as default according to ablation study as Section \ref{abs} described. % which is set to 0.7 as default. %and $\textbf{L}_\textbf{Teacher}$ are the loss functions as defined in equation (\ref{con:modelloss}). 

Recent studies \cite{tang2019distilling,lu2020twinbert,izacard2020distilling} has successfully adopted such kind of methods  to the area of monolingual IR. For example, \citeauthor{tang2019distilling}~\shortcite{tang2019distilling} and \citeauthor{lu2020twinbert}~\shortcite{lu2020twinbert} distill knowledge from the pre-trained language model BERT to a BiLSTM-based model and a non-interactive model, respectively. %builds non-interactive BERT-like encoders and distill. 
Contrast with these researches, we are the first to distill knowledge from a interactive model to the semi-interactive one, and examine the effectiveness of knowledge distillation under the cross-lingual scenario. Besides, the input of teacher model and student model are same in prior studies, while ours are different. 

%\newpage
\section{Experiment Setting}
\label{sec:exp}
We examine our methods on both semantic similarity task and document search task. 
In this section, we describe the experimental setting in detail. 
%We first train our models on \textbf{WikiCLIR} \cite{sasaki2018cross} and evaluate two tasks on both \textbf{WikiCLIR} and \textbf{CLIRMatrix} \cite{sun2020clirmatrix}..
%In this section, we examine the proposed methods on two CLIR datasets -- \textbf{WikiCLIR}  \cite{sasaki2018cross} and \textbf{MULTI-8} \cite{sun2020clirmatrix}.
%first introduce the dataset, the evaluation metrics, the experimental setting and the overall results. Then, ablation study of the proposed components are described.
\begin{table}
\begin{center}
\scalebox{0.9}{
\begin{tabular}{l|cccc}
\hline 
\multirow{2}*{\bf Direction} & \multicolumn{1}{c}{\bf Training}  & \multicolumn{1}{c}{\bf Validation}  & \multicolumn{2}{c}{\bf Testing}  \\ \cline{2-5}
& WikiCLIR  & WikiCLIR & \multicolumn{1}{c}{\begin{tabular}[c]{@{}c@{}}Semantic\\ Similarity\end{tabular}}  & \multicolumn{1}{c}{\begin{tabular}[c]{@{}c@{}}Document\\ Search\end{tabular}} \\ \cline{2-5}
\hline
Ru$\Rightarrow$En & 2.2M & 30.0K  & 60.0K & 100.0K  \\
Es$\Rightarrow$En & 2.8M & 30.0K  & 60.0K & 100.0K  \\
Fr$\Rightarrow$En & 5.0M & 30.0K  & 60.0K & 100.0K  \\
Pt$\Rightarrow$En & 1.6M & 30.0K  & 60.0K & - \\
Ar$\Rightarrow$En & 0.4M & 30.0K  & 60.0K & 100.0K  \\
De$\Rightarrow$En & - & -  & - & 100.0K  \\
Ja$\Rightarrow$En & - & -  & - & 100.0K  \\
Zh$\Rightarrow$En & - & -  & - & 100.0K  \\
\hline
Total & 12.0M & 150.0K  & 300.0K & 700.0K \\
\hline
\end{tabular}}
\end{center}
\caption{\label{tab:st1}Statistics of samples used in our study.  }
\end{table} 
%\subsection{Experimental Setting}
\subsection{Training and Validation Set}
\label{sec:data}
%Our training data compose of  \textbf{WikiCLIR} \cite{sasaki2018cross} and \textbf{CLIRMatrix} \cite{sun2020clirmatrix} datasets. 
%We construct our training  and validation set on \textbf{WikiCLIR} \cite{sasaki2018cross}.

% \begin{figure}[t]
% \centering
% \includegraphics[width=0.48\textwidth ]{LaTeX/picture/model/topk_mining_methods.pdf}
% \caption{An example of ready-made relevant queries collect method on \textbf{WikiCLIR} and \textbf{MULTI-8} .} 
% \label{fig:topkmethods}
% \end{figure}

\textbf{WikiCLIR} \cite{sasaki2018cross} is a large-scale CLIR dataset collected from Wikipedia which has provided positive and negative samples.  
We construct our training and validation set on  English (En) document and five multilingual queries in Russian (Ru), Spanish (Es), French (Fr), Portuguese (Pt) as well as Arabic (Ar). 
%For each language pair, we randomly extract K (K = thousand) and 6K(K = thousand) samples as the training and validation set, respectively. 
For each language pair, we randomly select 0.4M to 5.0M (M = million) samples as the training set, covering low-resource and high-resource language directions. And 60K (K = thousand) samples are extracted as the validation set for each language direction. The statistics of the used datset is concluded in Table~\ref{tab:st1}. 
% Different to \textbf{WikiCLIR} in which each query is labeled with only one document, \textbf{CLIRMatrix} 
% %reliably finds more relevant documents by propagating search results from monolingual IR systems to other languages via Wikidata and 
% allows for more finergrained levels of relevance, making the dataset suitable for the evaluation of retrieval tasks. 
Each document in \textbf{WikiCLIR} has been linked to multiple multilingual queries, we randomly collect $N$ ready-made relevant queries labeled as ``relevant'' for each document in the proposed semi-interactive mechanism. 
Finally, each sample is formed as:  ${<query,document,relevant\_queries,label>}$.
To cope with the pre-trained language model M-BERT, all the queries and documents are preprocessed following~\newcite{devlin2018bert},  the first token is a special classification token ([CLS]) and the document and relevant queries are concatenated with a special token ([SEP]).

\subsection{Evaluation Tasks and Metrics}
%We examine the proposed methods on two tasks. %NLP tasks and two  CLIR datasets.

%We examine the proposed methods on the following two tasks. % and a real-world search engine (described in Section \ref{as:real}).
%-- \textbf{WikiCLIR}  \cite{sasaki2018cross} and \textbf{MULTI-8} \cite{sun2020clirmatrix}.

\subsubsection{Semantic Similarity}
%This task is to measure the semantic equivalence between query and document. 
Semantic similarity tasks aim to predict whether query and document are semantically relevant or not.
%We use the \textbf{WikiCLIR} \cite{sasaki2018cross} dataset, a large-scale CLIR dataset collected from Wikipedia. Each English document is linked to multiple queries in different languages, associated with labels to represent whether each query-document pair is relevant. 
%Following the common setting \cite{sasaki2018cross}, we regard the first sentence or the title to represent the English (En) document. 
%Finally, each sample is  formed as : ${<query,document,label>}$.  
We evaluate our methods on \textbf{WikiCLIR} in five languages: Russian (Ru), Spanish (Es), French (Fr), Portuguese (Pt) as well as Arabic (Ar). For each language pair, we randomly extract 30K samples as the test set. 
%To cope with the pre-trained language model M-BERT, all the queries and documents are preprocessed following~\newcite{devlin2018bert}.
To evaluate the %semantic similarity task 
performance, we use the area under the curve  \citep[AUC,][]{1995Area}. 
%accuracy \citep[ACC,][]{diebold2002comparing}.
Moreover, in order to verify the effectiveness of mining queries from search log, we also conduct a group of experiments on a dataset collected from a real-world search engine, as described in Section \ref{as:real}.

\begin{table*}[t]
\begin{center}

\scalebox{1.0}{
\begin{tabular}{l|cccccc|c}
\hline 
\bf Model & \bf Ru$\Rightarrow$En  & \bf Es$\Rightarrow$En  & \bf Pt$\Rightarrow$En & \bf Fr$\Rightarrow$En & \bf Ar$\Rightarrow$En & \bf AVG & \bf Inf.Time(ms) \\ 
\hline
\textsc{Interact} & 90.71 & 96.57 & 96.56  & 96.37 & 90.86 & 94.86 &  45.8 \\ 
\hline
\textsc{Non-Interact} & 77.83  & 80.91 & 79.94   & 80.70  & 80.20  & 78.92 & 8.4  \\
\textsc{~~~~~+ KT} & 79.57  & 83.96  & 82.60  & 83.82  & 78.66  & 81.71  & 8.4 \\
\hline
\textsc{Semi-Interact} & \begin{math}82.31^{\ddagger}\end{math}  & \begin{math}85.81^{\ddagger}\end{math}  & \begin{math}84.98^{\ddagger}\end{math}   & \begin{math}85.68^{\ddagger}\end{math}  & \begin{math}84.44^{\ddagger}\end{math}  & \begin{math}84.04^{\ddagger}\end{math}  & 8.9 \\
\textsc{~~~~~+ KT} & \begin{math}\textbf{86.55}^{\ddagger}\end{math} &   \begin{math}\textbf{90.75}^{\ddagger}\end{math} &  \begin{math}\textbf{90.69}^{\ddagger}\end{math} &  \begin{math}\textbf{91.03}^{\ddagger}\end{math} &    \begin{math}\textbf{87.79}^{\ddagger}\end{math} &  \begin{math}\textbf{89.26}^{\ddagger}\end{math} &  8.9 \\
\hline
\end{tabular}}
\end{center}
\caption{\label{tab:main1} AUC of different models on WikiCLIR semantic similarity task. ``Inf. Time'' means the average inference time for each sample. $\ddagger$/$\dagger$ represents our model is significantly better than \textsc{Non-Interact} (p\textless0.01/0.05), tested by bootstrap resampling~\cite{Koehn2004Statistical}. }
\end{table*}

%补充cilir matrix数据集上的结果
\begin{table*}
\begin{center}
\scalebox{1.0}{
\begin{tabular}{l|cccc|ccc}
\hline 
\multirow{2}*{\bf Model}  &  \multicolumn{4}{c}{\bf Supervised} & \multicolumn{3}{c}{\bf Unsupervised}\\ \cline{2-8}
& {\bf Ru$\Rightarrow$En}  & {\bf Es$\Rightarrow$En}  & {\bf Fr$\Rightarrow$En} & {\bf Ar$\Rightarrow$En}  & {\bf De$\Rightarrow$En} & {\bf Zh$\Rightarrow$En} & {\bf Ja$\Rightarrow$En} \\ 

\hline
\textsc{\cite{sun2020clirmatrix}} & 0.7100  & 0.7600  & 0.7600  & 0.6000 & 0.7500 & 0.6300  & 0.7100 \\ 
\textsc{Interact} & 0.7717 & 0.7954 & 0.8050   & 0.7458 & 0.7579 & 0.7478  & 0.7504  \\ 
\hline
\textsc{Non-Interact} & 0.7052 & 0.6957  & 0.6965 & 0.7135 & 0.6669 & 0.7001  & 0.7062  \\
\textsc{~~~~~+ KT} & 0.7101 & 0.6989  & 0.6982 & 0.7203 & 0.6736 & 0.7024 & 0.7100  \\
\hline
%\textsc{Semi-Interact} & 0.7129 & 0.7131  & 0.7100 & 0.7315 & 0.6845 & 0.7049  & 0.7218
\textsc{Semi-Interact} & \begin{math}0.7129^{\ddagger}\end{math} &   \begin{math}0.7131^{\ddagger}\end{math} &  \begin{math}0.7100^{\ddagger}\end{math} &  \begin{math}0.7315^{\ddagger}\end{math} &    \begin{math}0.6845^{\ddagger}\end{math} &  \begin{math}0.7049^{\dagger}\end{math} &  \begin{math}0.7218^{\ddagger}\end{math} \\
%\textsc{~~~~~+ KT} & \bf 0.7259 & \bf 0.7181  & \bf 0.7255 & \bf 0.7328  & \bf 0.7041 & \bf 0.7252 & \bf 0.7419   \\
\textsc{~~~~~+ KT} & \begin{math}\textbf{0.7259}^{\ddagger}\end{math} &   \begin{math}\textbf{0.7181}^{\ddagger}\end{math} &  \begin{math}\textbf{0.7255}^{\ddagger}\end{math} &  \begin{math}\textbf{0.7328}^{\ddagger}\end{math} &    \begin{math}\textbf{0.7041}^{\ddagger}\end{math} &  \begin{math}\textbf{0.7252}^{\ddagger}\end{math} &  \begin{math}\textbf{0.7419}^{\ddagger}\end{math} \\
\hline
\end{tabular}}
\end{center}
%``\textsc{MT-Based}$^+$'' is trained using more than 20M En$\Rightarrow$En training samples in search log.  
%As seen, Our model significantly outperform baselines. }
\caption{\label{tab:main2} NDCG@10 of different models on MULTI-8 retrieval task. ``Supervised'' means the training data covers these retrieval directions, while ``Unsupervised'' does not.}% † is the same as defined in  Table~\ref{tab:main1}}
\end{table*}
\subsubsection{Document Search}
This task is to rank candidate documents by relevance with respect to a multilingual query.
%Considering the \textbf{WikiCLIR} only contains three kind of labels and the file formats correspond each query to only one docuemnt, it is neccary to find a more challenging dataset to evaluate our proposed methods.
Here, we use the test data of \textbf{MULTI-8} \cite{sun2020clirmatrix} to further evaluate our proposed methods.
 \textbf{MULTI-8} is also constructed from Wikipedia but allows for more finergrained levels of relevance. We evaluate our methods on seven multilingual queries (Arabic (Ar), German (De), Spanish (Es), French (Fr), Japanese (Ja), Russian (Ru), and Chinese (Zh)) with each language pair contains 100K samples.
%To evaluate the search performance, we use normalized discounted cumulative gain \citep[NDCG,][]{jarvelin2002cumulated}. 
%Following the default setting in \newcite{sun2020clirmatrix}, we calculate NDCG@10, which evaluates the top 10 recalled documents.
Following the default setting in \newcite{sun2020clirmatrix}, we calculate normalized discounted cumulative gain \citep[NDCG,][]{jarvelin2002cumulated} to evaluate the search performance.

\subsection{Implementation Details}
%The sentences in our data are tokenized and lowercased using the scripts provided in Moses.\footnote{\url{https://github.com/mosesdecoder}}. All the data are processed by byte-pair encoding to alleviate the Out-of-Vocabulary problem \cite{sennrich2015neural} with 40K merge operations. Finally, we extracted 70K vocabularies.\footnote{M-Bert uses word-piece to preprocess data and has its own vocabulary table. Accordingly, all the models built upon m-Bert are preprocessed following~\cite{devlin2018bert}.} 

We implement our model and the associated baselines upon Tensorflow. In order to preliminarily map different languages into the same latent space, all the encoders are initialized by the pre-trained cross-lingual language model {\bf M-Bert}~\cite{devlin2018bert}.\footnote{\url{https://github.com/google-research/bert}}  %that pre-trained as described in Section~\ref{sec:lm}.  
%The character, word and script vocabulary size are 13.5K, 58.4K and 107, respectively.
We follow \newcite{lu2020twinbert} to build each model with 6 layers in encoders. 
For the model training, we use Adam optimizer \cite{DBLP:journals/corr/KingmaB14} and set learning rate to 3e-5 with 8K warmup steps. 
Besides, dropout rate is set to a constant of 0.1 to enhance the model robustness. 
We train our model on a single Tesla P100 GPU with a mini-batch consisting of 1,024 samples.  
The training of each model is early-stopped to maximize AUC on the validation set.  Other hyper-parameters are assigned with the common finetune configurations described in \newcite{devlin2018bert}. We compare following models:

\begin{itemize}
\item \textsc{Interact}: We first build an interactive model as baseline and  teacher model. Since there are few studies explore V-CLIR, we build the model following \newcite{jiang2020cross}. Contrary to previous study, we initialize the encoder using multilingual BERT rather than the monolingual version~\cite{devlin2018bert}. 
\item \textsc{Non-Interact}: 
We borrow the advanced monolingual vector-based IR model  \cite{reimers2019sentence} into the CLIR task as our non-interactive baseline. 
\item \textsc{Ours}: We examine the proposed semi-interactive mechanism (\textsc{Semi-Interact}) and knowledge transfer method (\textsc{KT}). Considering the former, we randomly select $N=3$ relevant queries for each document. For the latter, we set $\alpha$ to 0.7 as default.  % There are some methods to close the gap between non-interactive and interactive architecture, that is to use knowledge distillation\cite{sanh2019distilbert,sun2019patient}.
% cross-lingual The m-BERT model's parameters were randomly initialized, and was applied to get russian query embedding and english item embedding.
% \item \textsc{MT-Based}:  For the most widely used method, we also compare our approach with machine translation based model. We translate multilingual queries to English, and call the En$\Rightarrow$En retrieval of the well-trained \textsc{M-Bert-Non}. We use the Ru$\Rightarrow$En and Es$\Rightarrow$En machine translation models that publicly available in an open source project Tatoeba~\footnote{\url{https://github.com/Helsinki-NLP/Tatoeba-Challenge}}~\cite{TiedemannThottingal:2020}. These models are trained using around 100M parallel sentences, making our evaluation convinced. %  the released models At first, the query was translated from russian to english by Google-Translation. And then, BERT was applied to get english query embedding and item embedding.
% %\item m-BERT$_{ru2en}$: The m-BERT model directly was applied to get russian query embedding and english item embedding. 
\end{itemize}

% \begin{figure*}[t]
% \centering
% \subfigure[Ru$\Rightarrow$En]
%     {
%         \begin{minipage}[t]{0.32\textwidth}
%         \centering         %子图居中
%         \includegraphics[width=1\textwidth ]{LaTeX/picture/model/AAAI21_ndcg1.pdf}
%         \end{minipage}%
%     }
% \subfigure[Es$\Rightarrow$En]
%     {
%         \begin{minipage}[t]{0.32\textwidth}
%         \centering         %子图居中
%         \includegraphics[width=0.9\textwidth]{LaTeX/picture/model/AAAI21_ndcg2.pdf}
%         \end{minipage}%
%     }
% \subfigure[En$\Rightarrow$En]
%     {
%         \begin{minipage}[t]{0.32\textwidth}
%         \flushleft         %子图居中
%         \includegraphics[width=0.9\textwidth]{LaTeX/picture/model/AAAI21_ndcg3.pdf}
%         \end{minipage}%
%     }
% \caption{nDCGs of different CLIR models. Our model is significantly better than baseline in all the  retrieval directions.  }
% \label{fig:nDCGs}
% \end{figure*}

\section{Results}
\subsection{Main Results}
%\subsubsection{Results on Relevance Task}
\subsubsection{Results on Semantic Similarity}

%We first compare our model with baselines. %In this serious of experiments, all the data are used to train a unified model.
%In the relevance and retrieval experiment, machine translation based BERT and m-BERT were selected as the baselines and compared with our  EComLM. ALL models were trained on the same training data with the same hyperparameters as described in the relevance experiment. 
%\paragraph{Relevance Task}
As shown in Table~\ref{tab:main1}, our model significantly outperforms the
non-interactive baseline across different language pairs, demonstrating the universal-effectiveness of the proposed method. In the meanwhile, our model yields over 4 times faster than its interactive counterpart at the inference time. %These results indicating the effectiveness of the proposed approaches.

Concretely, although interactive model outperforms non-interactive model, it significantly increases the inference time and fails to be employed in a real-world IR system. We regard this model as an upper limit of retrieval quality. The proposed semi-interactive mechanism narrows the accuracy gap, which confirms our hypothesis that supplementing the document representation with its relevant multilingual contexts is conductive to V-CLIR. Moreover, the interactions among multilingual queries and documents of our approach are able to be conducted offline, maintaining the superiority of non-interactive model on processing speed. 

We assess knowledge transferring on non- and semi-interactive models. Both the models boost the retrieval performance, indicating that transferring knowledge from a well-trained interactive model to non-interactive models is able to narrow the quality gaps among them. Besides, our results demonstrate that the \textsc{Semi-Interact} and \textsc{KT} are complementary to each other, progressively benefiting to the cross-lingual semantic matching.

%\subsubsection{Results on Retrieval Task}
\subsubsection{Results on Document Search} 
%We further examine our method in the retrieval task. 
Table~\ref{tab:main2} shows the NDCG@10 of different models on MULTI-8 document search tasks. 
Similar to semantic similarity tasks, our method yields consistently better results across language pairs. As a superiority of pre-trained multilingual language model lies in its ability on knowledge transferring, we can transfer knowledge from high-resource languages  to low-resource ones in the downstream tasks. Accordingly, we conduct three retrieval tasks that are not considered at the training time.  As seen, the proposed methods yield considerable performance on both the three tasks. It is encouraging to see that, although we do not enhance documents with relevant queries in these three languages, the improvements are still preserved on these tasks.   

\subsection{Ablation Study}
\label{abs}
In this section, we conduct ablation study of different components on \textbf{WikiCLIR} semantic similarity tasks. We report the average score for simplification. 
\subsubsection{Effects of the Number of Relevant Queries} 
\begin{figure}[t]  %[htbp]中的h是浮动的意思
    \centering
    \subfigure[Number $N$ of Queries]
    {
        \begin{minipage}[t]{0.21\textwidth}
        \centering         %子图居中
        \includegraphics[width=1.0\textwidth ]{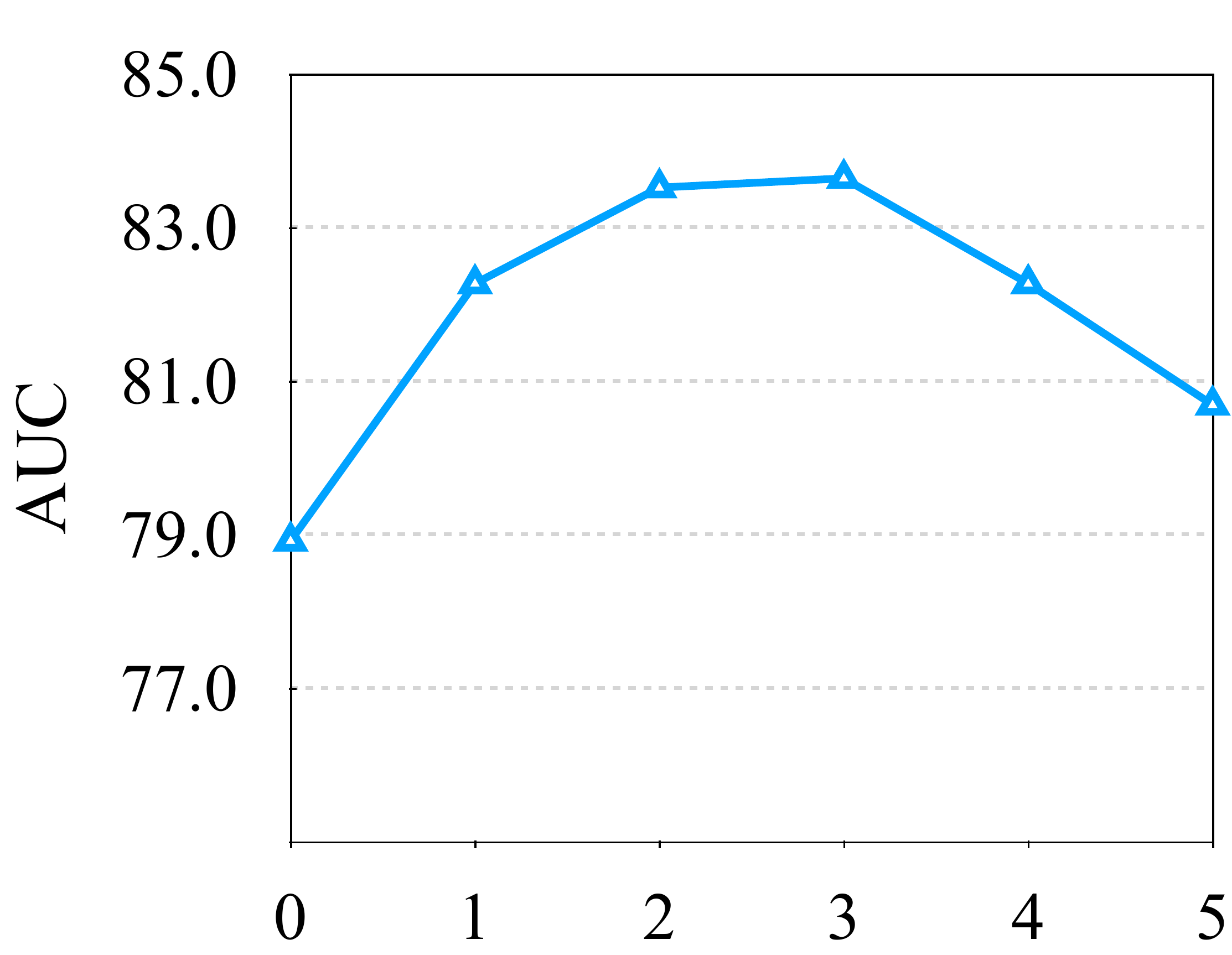}
        \end{minipage}%
    }~~~~~~
    \subfigure[Factor $\alpha$ in Loss]
    {
        \begin{minipage}[t]{0.21\textwidth}
        \centering         %子图居中
        \includegraphics[width=1.0\textwidth ]{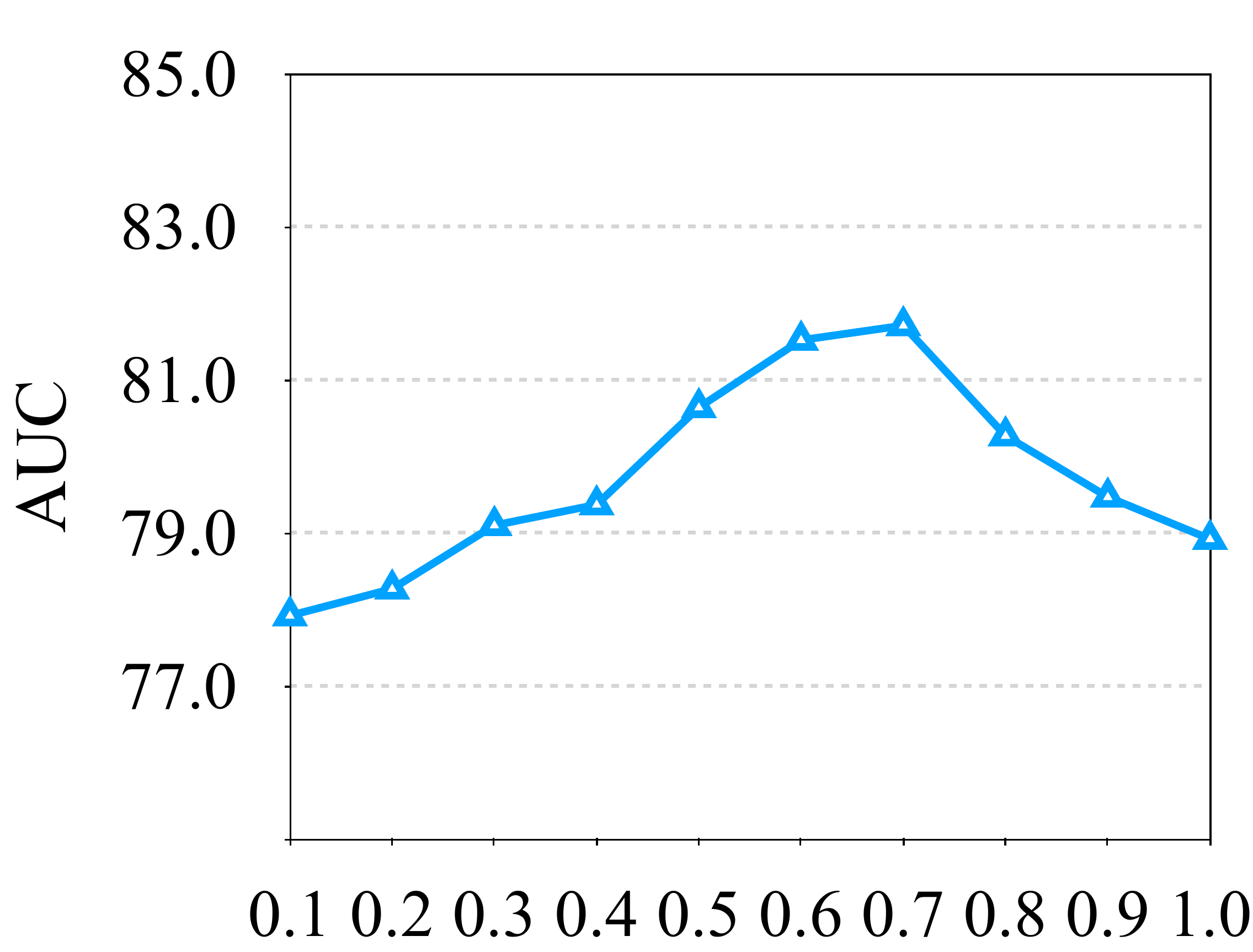}
        \end{minipage}%
    }
    %\includegraphics[width=0.3\textwidth]{LaTeX/picture/model/sigir_topk.pdf} % Reduce the figure size so that it is slightly narrower than the column.
    %ECom\_LM
    \caption{Effects of (a) the number of relevant queries being used in document encoding, and (b) the value of weights $\alpha$ in loss. Experiments are conducted based on \textsc{Semi-Interact}.} %a. a = 0.7 yields the best result. . K = 5 yields the best result. } % 
    \label{fig:k} 
\end{figure}

We plot Figure~\ref{fig:k} (a) to show the effects of the number of relevant queries added to document representation. Specifically, via incorporating queries (N $>$ 0), the relevance accuracy increases, confirming that the associated queries are helpful for the document representation learning. We observe that the number with 3 is superior to other settings. When the number of  queries goes up (N $>$ 3), the classification performance inversely drops. One possible reason is that the conventional semantic meaning of the document may be potentially overlooked when 
complementing superfluous multilingual queries. This results in biases in document representations and raises the difficulty on relevance prediction. % potentially introduce biases into the document representation. The conventional semantic meaning of the document may be %  marginally contribute to our model, but also

%By merging strategy 1 and strategy 2, we get the best results
%To test the inference time, we ran benchmarks on a workstation with the following configuration:  GPU V100, CPU 64 core and 96G memory. To eliminate the influence of noise on queries, we evaluated the average inference time on 100,000 queries (the average length of query word piece is 6, the average length of title is 28), and the results are summarized in Table 7.

% \subsubsection{Effects of Document Length}

% ----------------

% ----------------

% ----------------

% ----------------

% ----------------

% ----------------

% ----------------

% ----------------

% ----------------

% \usepackage{multirow}
\begin{table}[]
\centering
\scalebox{1.0}{
\begin{tabular}{lccc}
\hline
\multicolumn{2}{c}{\multirow{2}{*}{\bf Model}}        & \multicolumn{2}{c}{\textbf{Training}} \\ \cline{3-4} 
\multicolumn{2}{c}{}                                       & {\Checkmark}                 & { \XSolidBrush }                 \\ \hline
\multicolumn{1}{c|}{\multirow{2}{*}{\textbf{Testing}}} & {\Checkmark}  & 84.04 & 78.76              \\
\multicolumn{1}{c|}{}                                  & { \XSolidBrush }  & 79.97 & 78.92                  \\ \hline
\end{tabular}
}
\caption{\label{tab:erqu} Ablation studies on whether relevant queries are incorporated during training and testing.} %\Checkmark and \XSolidBrush respectively represent whether to use relevant queries during training and testing .}
\end{table} 
\subsubsection{Usage of Semi-Interactive Mechanism} 
We verify the effect of semi-interactive on different stages, i.e. training and inference.  
%We verify the effectiveness of the relevant queries by whether the relevant queries are used during training and testing. 
As shown in Table \ref{tab:erqu}, we found that exploiting relevant queries at the training time can consistently improve the performance, no matter relevant queries are utilized or not during inference. %but not during testing, the results have also been improved. 
When adapting semi-interactive mechanism only in inference, the performance marginally changes. 
We attribute these to the fact that complementing multilingual contexts during training benefits the representation learning. Our results suggest that  carrying out our method on both training and inference stages performs best. 
%At the same time, if related queries are only used during testing, it will have a negative impact on the results.
%This shows that the relevant queries are indeed useful, but the premise is to change the model structure to be effective. 
\subsubsection{Effects of Factor in Distillation} Figure~\ref{fig:k} (b) visuals the effects of the factor $\alpha$ used in loss function at knowledge distilling time. Our results reveal that the weight significantly affect the performance. A small $\alpha$ makes the teacher model plays a dominate role during training time, while too large values result in less affects from the distillation procedure. Weight with 0.7 is the best configuration to balance the learning from the groundtruth and the teacher model. %When the weight is too small or too large, the performance is not good. We guess that the role of the teacher model is to guide training, not to dominate.
\iffalse
\begin{figure}[t]  %[htbp]中的h是浮动的意思
    \centering
    \includegraphics[width=0.3\textwidth]{LaTeX/picture/model/sigir_kda.pdf} % Reduce the figure size so that it is slightly narrower than the column.
    %ECom\_LM
    \caption{Effects of the value of loss function weights a. a = 0.7 yields the best result. } % 
    \label{fig:loss} 
\end{figure}
\fi

\subsubsection{Effectiveness of Knowledge Transferring Strategies} We further conduct experiments to examine knowledge transferring strategies. As shown in Table~\ref{tab:kd}, reusing teacher model word embeddings (\textsc{Initialize}) and knowledge distillation (\textsc{Distillation}) improve the relevance quality separately and are complementary to each other. These results verify again that transferring knowledge from interactive model is facilitate to the representation learning and relevance scoring of non-interactive model.  %superimposed effect of different knowledge distillation strategies.By merging two strategies, modifying the loss function((\textsc{M-Bert-Semi-Kd-1})) and initializing language model embedding((\textsc{M-Bert-Semi-Kd-2})), we get the superimposed knowledge distillation results.

\begin{table}
\centering
\scalebox{1.0}{
\begin{tabular}{lc}
\hline 
\bf Model %&\multicolumn{2}{c}{\bf AVG} \\ \cline{2-3}
& AUC \\ %\cline{2-3}
\hline
\textsc{Semi-Interact}  & 84.04   \\
\textsc{~~~~+ Initialize} &  88.33  \\
\textsc{~~~~+ Distillation} & 87.54  \\
%\textsc{Semi-Interact + KT} & \bf 89.26 & \bf 80.21 \\
\textsc{~~~~+ Initialize~~+ Distillation} & \bf 89.26 \\

\hline
\end{tabular}
%By merging two strategies, we get the superimposed knowledge distillation results.}
}
\caption{\label{tab:kd} Ablation studies on knowledge transferring.}
\end{table}

% \begin{table}
% \scalebox{1.0}{
% \begin{tabular}{lcc}
% \hline 
% \bf Model %&\multicolumn{2}{c}{\bf AVG} \\ \cline{2-3}
% & AUC & ACC\\ %\cline{2-3}
% \hline
% \textsc{Semi-Interact}  & 84.04 & 78.69  \\
% \textsc{~~~~+ Initialize} &  88.33 &  79.51 \\
% \textsc{~~~~+ Distillation} & 87.54 & 79.84 \\
% %\textsc{Semi-Interact + KT} & \bf 89.26 & \bf 80.21 \\
% \textsc{~~~~+ Initialize~~+ Distillation} & \bf 89.26 & \bf 80.21 \\

% \hline
% \end{tabular}
% %By merging two strategies, we get the superimposed knowledge distillation results.}
% }
% \caption{\label{tab:kd} Ablation studies on knowledge transferring strategies.}
% \end{table} 

% 向量可视化结果对比（待补充by xll）
%\section{Analysis} 
\subsection{Representation Visualization}
A natural question is why and how the proposed mechanism boosts the model performance. We pay our attention into the representation learning of queries and documents. 
%Pre-trained models such as BERT have achieved great success in many natural language processing tasks. However, how to obtain better sentence representation through these pre-trained models is still worthy to exploit. 
\citeauthor{DBLP:journals/corr/abs-2103-15316}~\shortcite{DBLP:journals/corr/abs-2103-15316} and \citeauthor{DBLP:conf/iclr/GaoHTQWL19}~\shortcite{DBLP:conf/iclr/GaoHTQWL19} pointed out that sentence embeddings not in a standard orthogonal basis cause poor performance of semantic similarity tasks, especially for the BERT-like models. %may caused by being not in a standard orthogonal basis. 
%\cite{,DBLP:conf/iclr/GaoHTQWL19} explore the reason for the poor performance of BERT-based sentence embedding in similarity matching tasks, i.e., it is not in a standard orthogonal basis.

\begin{figure}[t]
\centering
\subfigure[Interactive ]
    {
        \begin{minipage}[t]{0.2\textwidth}
        \centering         %子图居中
        \includegraphics[width=0.8\textwidth ]{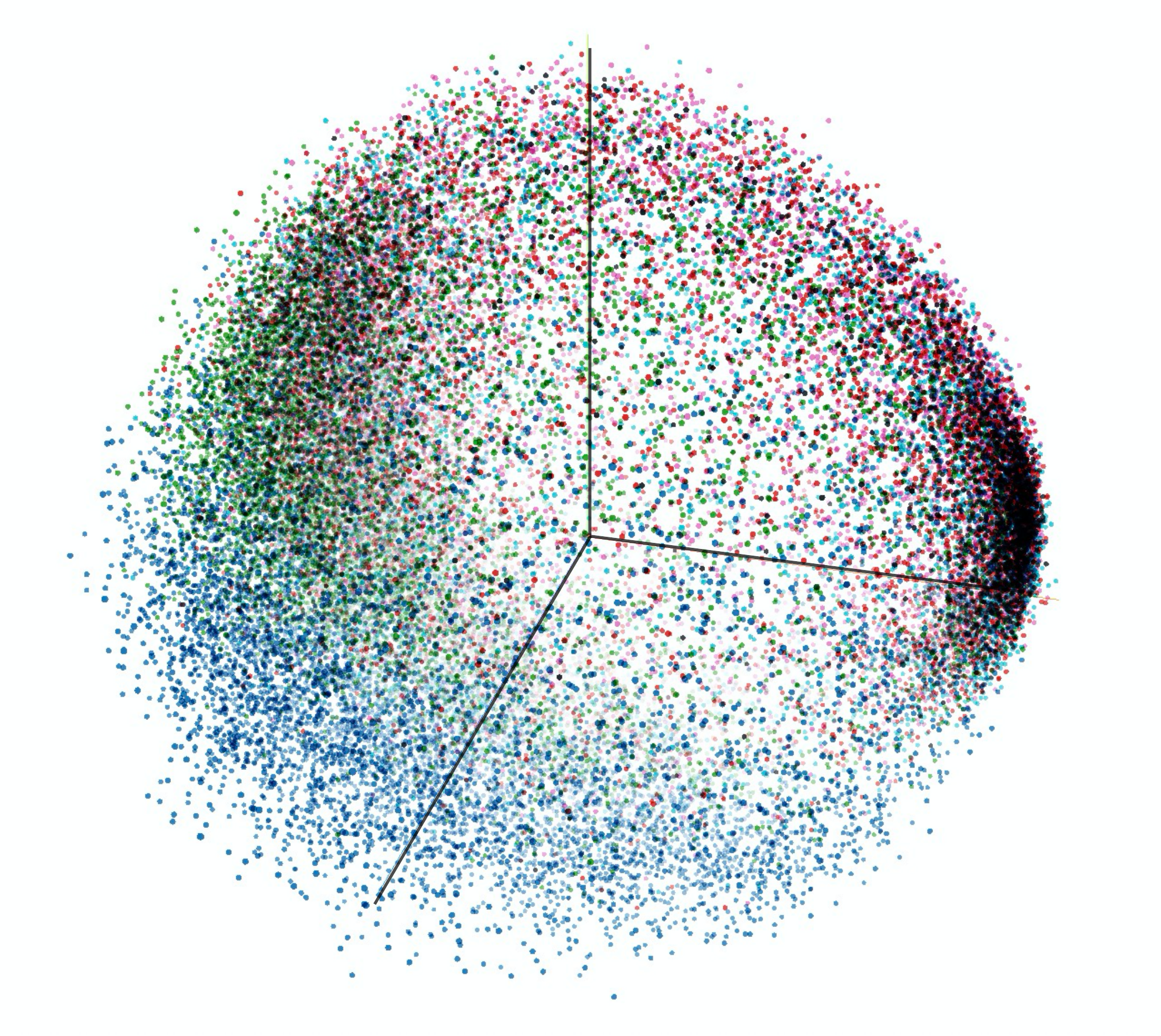}
        \end{minipage}%
    }
\subfigure[Non-Interactive ]
    {
        \begin{minipage}[t]{0.2\textwidth}
        \centering         %子图居中
        \includegraphics[width=0.8\textwidth]{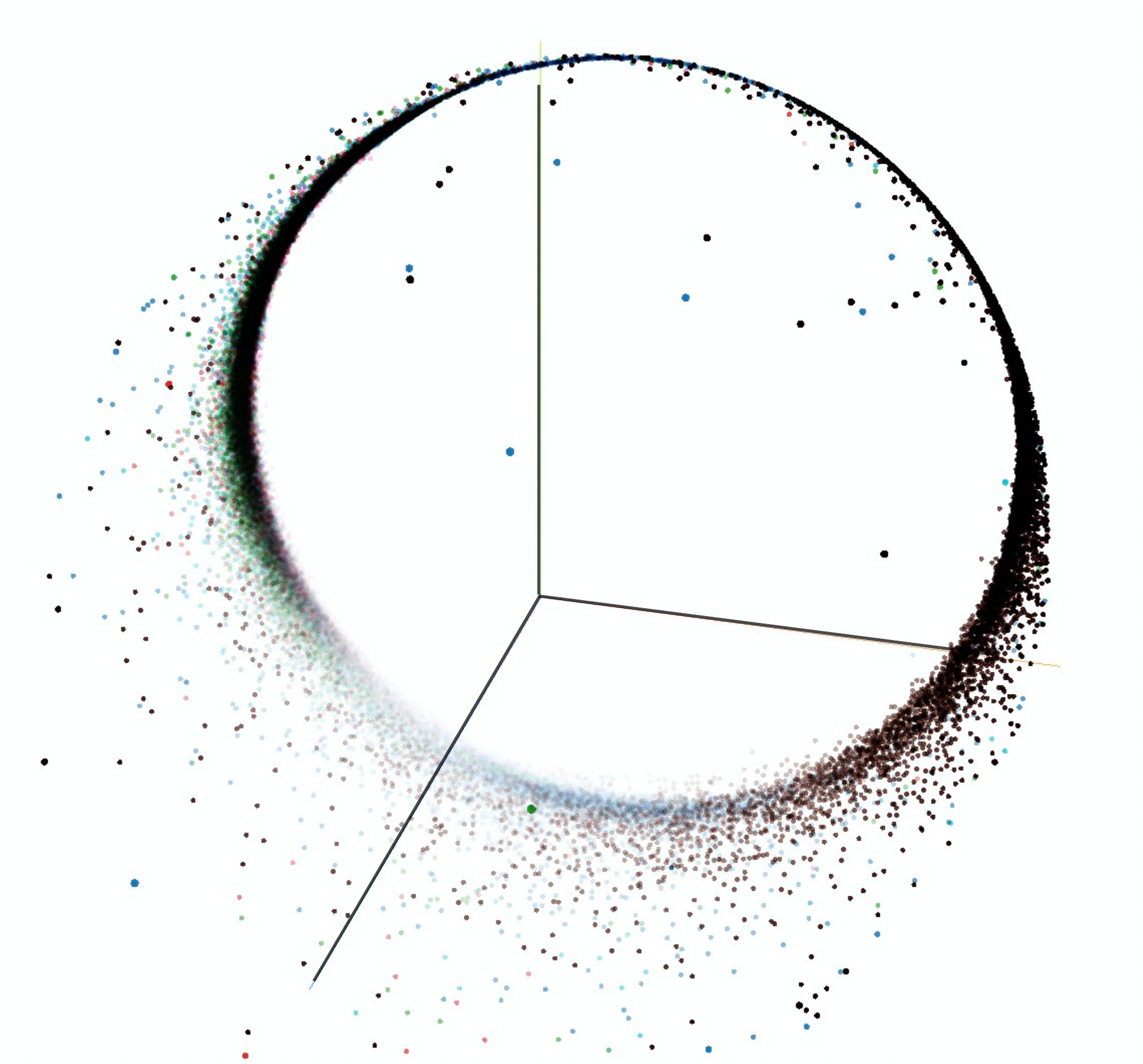}
        \end{minipage}%
    }
\subfigure[Semi-Interactive]
    {
        \begin{minipage}[t]{0.2\textwidth}
        \centering          %子图居中
        \includegraphics[width=0.7\textwidth]{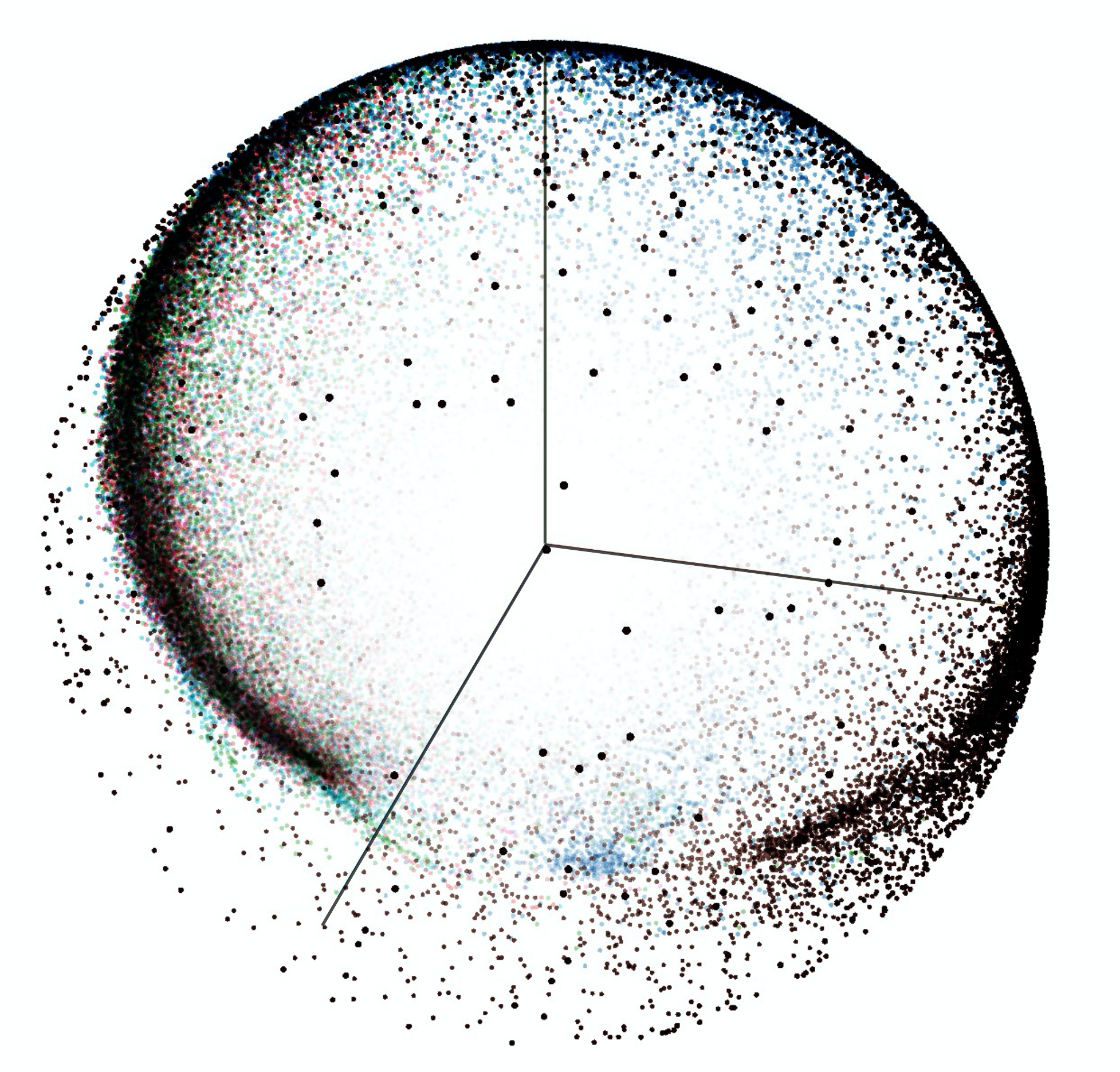}
        \end{minipage}%
    }
\subfigure[Semi-Interactive + KT]
    {
        \begin{minipage}[t]{0.2\textwidth}
        \centering           %子图居中
        \includegraphics[width=0.72\textwidth]{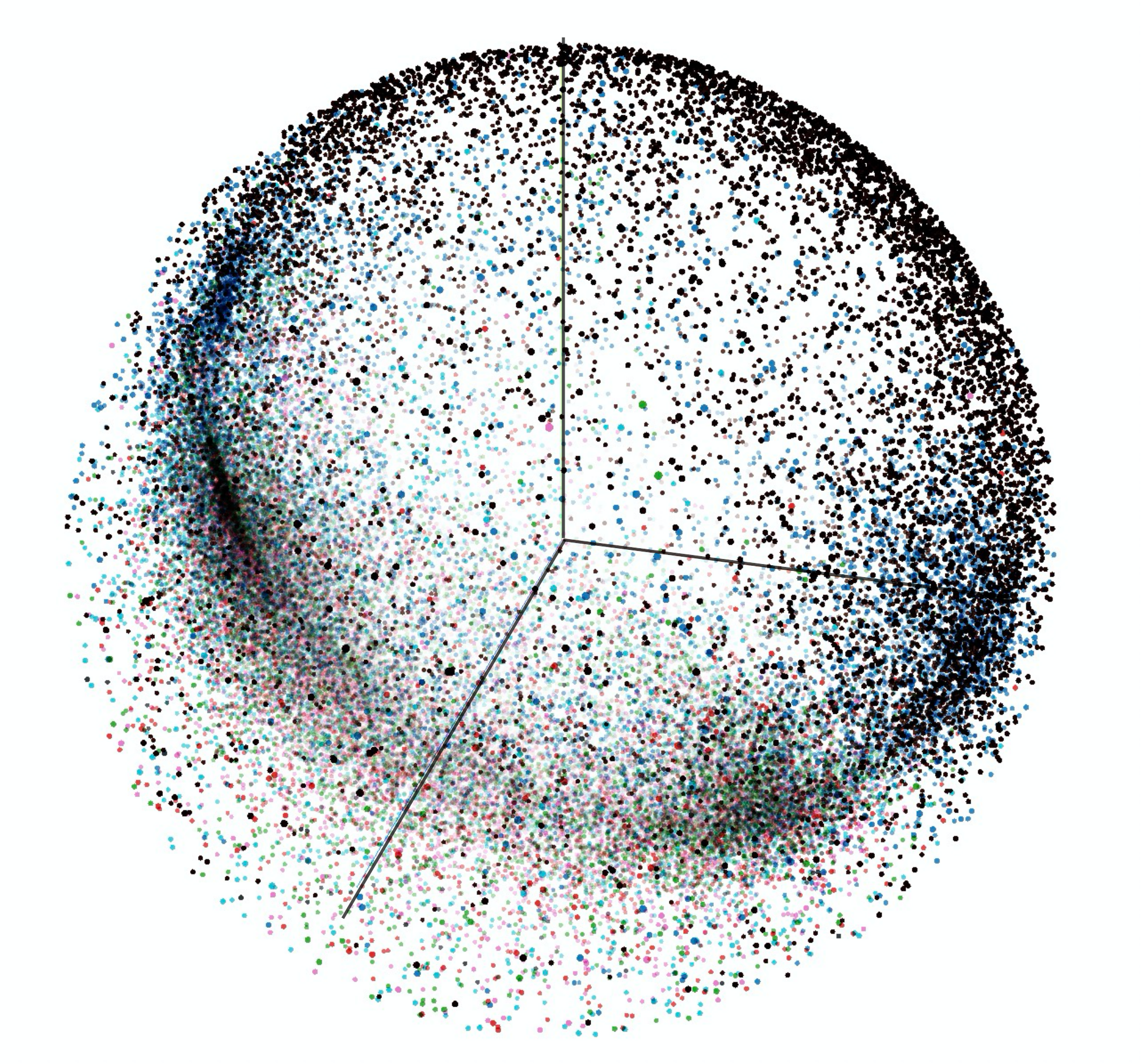}
        \end{minipage}%
    }
\caption{3D visualization of query and document representations distribution. 
%Each point in the figure represents a query or document representation, and each color represents a document or a different type of query.
Each point in the figure represents a query or document representation. Different languages are distinguished by different colors.
%a) Visualization of query and document representation trained from interactive mechanism, b) Visualization of query and document representation trained from non-interactive architecture , c) Visualization of query and document representation trained from semi-interactive mechanism , and d) Visualization of query and document representation trained from  the proposed semi-interactive mechanism + KT . 
Compared with non-interactive, the representations of semi-interactive are gradually distributed to more uniform representation space. 
} 
\label{fig:view}
\end{figure}

For a more intuitive comparison, we randomly select 10K query and document representations from our training set, and project them into a 3-dimensional space using principal component analysis \citep[PCA,][]{abdi2010principal}. 
%project the learned query and document representations into a 3-dimensional space using principal component analysis \citep[PCA,][]{abdi2010principal} for visualization on WikiCLIR.  
As shown in Figure~\ref{fig:view}, the representations learnt from interactive model (Figure~\ref{fig:view}(a)) are diversely distributed around the origin using PCA projection.
In contrast, the representations in non-interactive model (Figure~\ref{fig:view}(b)) degenerated into a narrow boundary, which means that the representations do not have enough capacity to model the diverse semantics in natural languages~\cite{yang2017breaking,mccann2017learned}.
Furthermore, we find the representations of semi-interactive (Figure~\ref{fig:view}(c)) are gradually distributed to the origin, which is more obviously while applying transferring knowledge methods (Figure~\ref{fig:view}(d)).

% 待调整

\subsection{Mining Relevant Queries from Search Log}
\label{as:real}
All the above experiments are based on ready-made queries provided in Wiki pages. We are interested at the effectiveness of the proposed semi-interactive mechanism in a real-world search engine where off-the-rack multilingual relevant queries for each document are not available.  

To this end, we examine our method on our in-house dataset extracted from a real-world search engine Alibaba.com.
\footnote{Note that, we cannot release the data, we report this results to illustrate the effectiveness of our approach in a real-world scenario. Readers can re-produce our results using the open-released dataset described in Section \ref{sec:data}.} % For anonymity, we temporarily use Alibaba.com to indicate the name of this real-world cross-lingual search engine. } 
Specifically, we conduct experiments on Russian-to-English and Spanish-to-English
%, French-to-English, Portuguese-to-English, and Arabic-to-English 
semantic similarity tasks. Query-title training pairs are collected from the search logs, resulting in 1 million training samples for each task.
%The relevance is labeled according to click behavior of users. 
For each query, we randomly sample documents from the document pool as negatives, while using the one user clicked as positive.
The test set is extracted from training data and manually checked by bilingual experts, resulting in 30K samples for each directions. We select top 3 multi-lingual queries for each document according to clickthrough data (as described in Section \ref{sec:topkm})).
%We examined the proposed approaches upon our in-house pre-trained model. 

As shown in Table~\ref{tab:semi}, our model significantly outperforms non-interactive based approach over 7 AUC in average. The results confirm the effectiveness of our method when mining relevant queries from search log. 
%In addition, ABTest demonstrates that our model improves clickthrough rate (CTR) by more than 5\%.
%In addition, we evaluate our model on real-world search with significant metrics gains observed in online A/B experiments.
 
\begin{table}[t]
\begin{center}
\begin{tabular}{l|c c|c}
\hline \bf Model & \bf Ru$\Rightarrow$En  & \bf Es$\Rightarrow$En  & \bf AVG  \\ \hline
\textsc{Interact}  & 84.96 &85.67 & 85.38   \\ \hline
\textsc{Non-Interact} & 68.54 &69.83 & 69.65  \\
\textsc{~~~~~+ KT} & 69.90 &70.64 & 70.39   \\ \hline
\textsc{Semi-Interact} & 72.97 & 73.42 &  73.21 \\ 
\textsc{~~~~~+ KT} & \textbf{75.66} & \textbf{77.63} & \textbf{76.90} \\ \hline
% \textsc{EComLM-Non}   &  70.31   \\
% \textsc{EComLM-Semi}   & 72.83 \\
\end{tabular}
\end{center}
\caption{\label{tab:semi} AUC of models when mining relevant queries from a real-world search log.  }
\end{table}

\section{Conclusion}
\label{sec:length}
In this paper, we provide insights on how to leverage the advantages of interactive model and non-interactive model for vector-based CLIR. 
Via supplementing multilingual features into document representation and transferring knowledge from a well-trained interactive model, the proposed model yields significant improvements over the conventional non-interactive model, while preserving its advantage with respect to computational efficiency. Our in-depth analyses further suggest that 1) the proposed strategies can tackle the representation degeneration problem in non-interactive models, thus obtaining discriminative representations; and 2) both of ready-made queries and mined queries are effect for supplementing multilingual features to documents. 

Several directions are worth to be explored in future. 
%\begin{itemize}
%    \item 
    %In this study, we simply exploit the relevant multilingual queries for each document provided in the examined dataset. A potential direction in future is to 
    %We explore the collection of multilingual queries under a real-world scenario, e.g. 
    %collecting top queries according to clickthrough data \cite{radlinski2008does,ma2008learning}, 
An interesting problem is to explore more effective approaches to collect  multilingual queries, especially for few-shot tasks, e.g. generating associated multilingual queries using text summarization \cite{gambhir2017recent,liu2019text}. 
%or query expansion \cite{carpineto2012survey,kuzi2016query} techniques. 
Another potential direction is to select queries with a more flexible manner such as using a dynamic routing layer  \cite{sabour2017dynamic,li2019information}.

\bibliography{Formatting-Instructions-LaTeX-2022}

\end{document}